\documentclass[10pt,twocolumn,letterpaper]{article}

\usepackage[pagenumbers]{wacv} 

\usepackage{graphicx}
\usepackage{amsmath}
\usepackage{amssymb}
\usepackage{booktabs}
\usepackage{diagbox}

\usepackage[accsupp]{axessibility}  

\usepackage[pagebackref,breaklinks,colorlinks]{hyperref}

\newcommand{\comment}[1]{} 

\usepackage{xcolor}
\definecolor{patrick_color}{rgb}{.0,.6,.05}
\definecolor{chengcheng_color}{rgb}{.5,.7,.1}
\definecolor{chris_color}{rgb}{0,0.35,0}
\definecolor{jeremy_color}{rgb}{0,0,1}
\definecolor{charlie_color}{rgb}{0,0,0.8}
\definecolor{james_color}{rgb}{0.75,0.25,0.0}
\definecolor{kunal_color}{rgb}{0.6,0.0,0.6}

\newcommand{\networkName}{PressureVision++\xspace}
\newcommand{\methodName}{PressureVision++\xspace}
\newcommand{\datasetName}{ContactLabelDB\xspace}

\usepackage[capitalize]{cleveref}
\crefname{section}{Sec.}{Secs.}
\Crefname{section}{Section}{Sections}
\Crefname{table}{Table}{Tables}
\crefname{table}{Tab.}{Tabs.}

\def\wacvPaperID{1133} 
\def\confName{WACV}
\def\confYear{2024}

\begin{document}


\title{PressureVision++: Estimating Fingertip Pressure from Diverse RGB Images} 

\author{Patrick Grady\textsuperscript{1}, Jeremy A. Collins\textsuperscript{1}, Chengcheng Tang\textsuperscript{2}, Christopher D. Twigg\textsuperscript{2},\\Kunal Aneja\textsuperscript{1}, James Hays\textsuperscript{1}, Charles C. Kemp\textsuperscript{1}\\
\\
\textsuperscript{1}Georgia Institute of Technology,~\textsuperscript{2}Meta Reality Labs\\
}

\twocolumn[{%
\renewcommand\twocolumn[1][]{#1}%
\maketitle

\begin{center}
\centering
\vspace{-7mm}
\includegraphics[width=16.9cm]{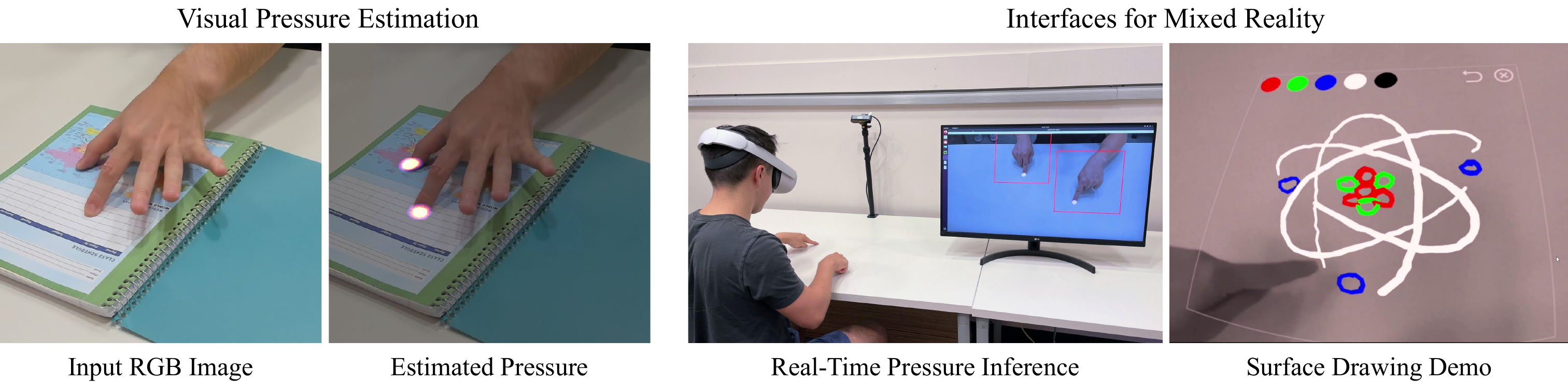}
\end{center}
\vspace{-2mm}
{\small Figure 1. \networkName{} leverages \emph{contact labels} to estimate fingertip pressure during interaction with diverse surfaces. Our method enables mixed reality devices to use everyday surfaces as touch-sensitive interfaces, as demonstrated with a surface drawing application.}
\vspace{0mm}
\label{fig:headliner}
\vspace{4mm}
}]
\setcounter{figure}{1}  

\begin{abstract}
Touch plays a fundamental role in manipulation for humans; however, machine perception of contact and pressure typically requires invasive sensors. Recent research has shown that deep models can estimate hand pressure based on a single RGB image. However, evaluations have been limited to controlled settings since collecting diverse data with ground-truth pressure measurements is difficult. We present a novel approach that enables diverse data to be captured with only an RGB camera and a cooperative participant. Our key insight is that people can be prompted to apply pressure in a certain way, and this prompt can serve as a weak label to supervise models to perform well under varied conditions. We collect a novel dataset with 51 participants making fingertip contact with diverse objects. Our network, \networkName{}, outperforms human annotators and prior work. We also demonstrate an application of \methodName{} to mixed reality where pressure estimation allows everyday surfaces to be used as arbitrary touch-sensitive interfaces. Code, data, and models are available online.\footnote{\url{https://pressurevision.github.io/}}
\end{abstract}


\section{Introduction} \label{sec:intro}
People frequently interact with their surroundings by applying pressure with their hands.  Machine perception of hand contact pressure has been used for activity recognition \cite{DBLP:journals/nature/SundaramKLZ0M19}, ergonomics \cite{reinvee2014utilisation}, user interfaces \cite{roh2011robust}, and other applications. Most approaches use physical pressure sensing arrays. These sensors, however, may be expensive or impractical to mount to hands or natural objects.


Recently, PressureVision \cite{grady2022pressurevision} showed that computer vision can be used to estimate hand pressure from a single RGB image. As opposed to physical pressure sensors, cameras provide a scalable, low-cost method to sense contact and pressure, opening the door to broad application.
While their model performs well with diverse hands, performance was only explored in an idealized, controlled environment. Their capture environment used artificial lighting, high-end machine vision cameras, and was trained and tested on simple, flat, rigid surfaces. Based on their results, it is not evident that vision-based pressure estimation is possible in less constrained settings.

\begin{figure}
  \centering
  \includegraphics[width=0.9\linewidth]{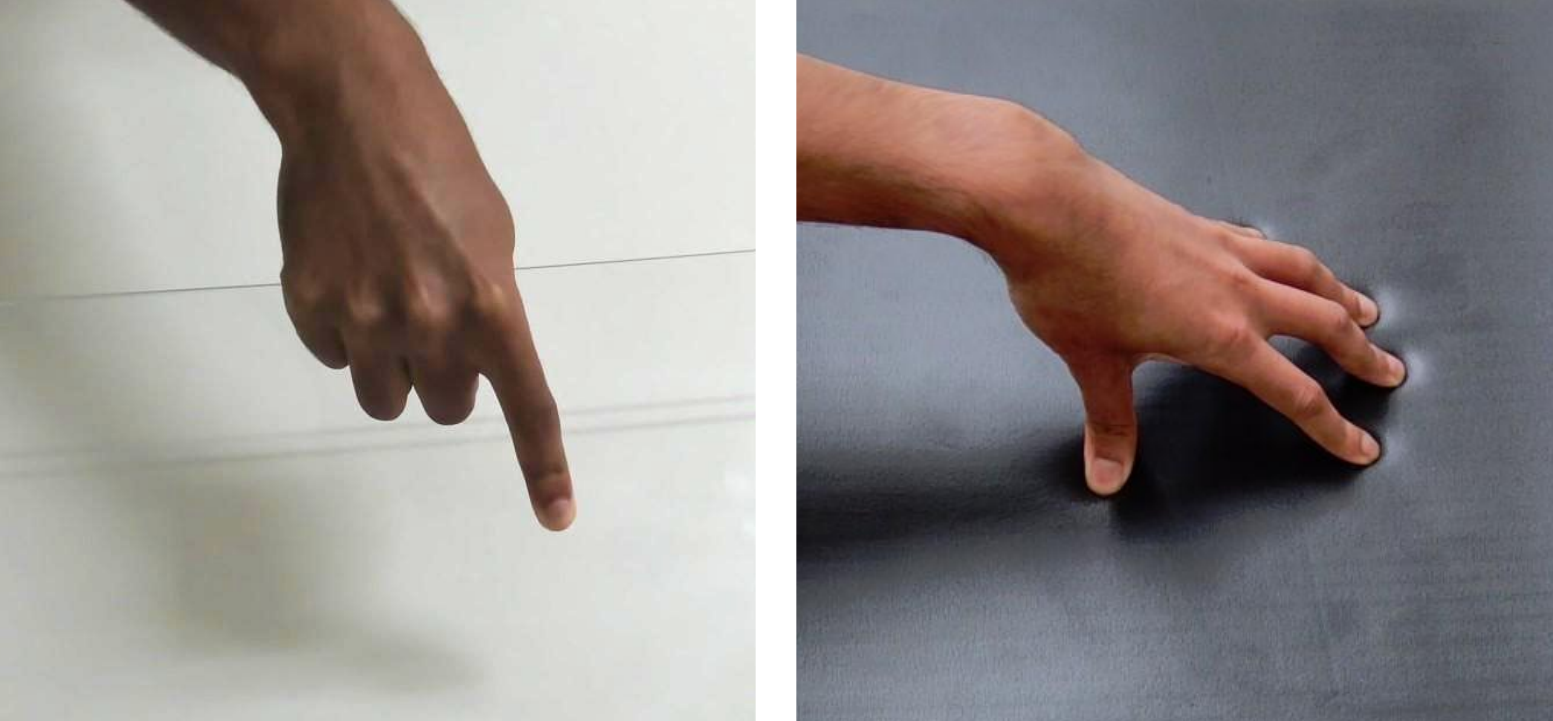}
  \caption{Instrumenting surfaces with pressure sensors without altering their properties is challenging. For example, pressure sensors must be transparent in order to instrument glass, and must be stretchable in order to instrument a deformable mat.}
  \label{fig:data_cap}
\end{figure}

Collecting training data for diverse surfaces might enable PressureVision \cite{grady2022pressurevision} to generalize more broadly, but data collection is challenging (Figure \ref{fig:data_cap}). Each RGB image in the training data requires ground truth pressure from a high-resolution pressure sensor. This requirement severely limits data collection for diverse surfaces, since mounting pressure sensors alters the appearance and properties of the surface. Additionally, human labelers have difficulty identifying contact and pressure from images.

We present a novel approach that enables training data for visual hand pressure estimation to be captured for unaltered surfaces found in the wild. Instead of instrumenting the surface, our approach relies on people's manual dexterity and highly sensitive perception of touch~\cite{touch}. We prompt participants to make contact with a surface in a specific way or move their hands close to a surface without making contact. The prompts serve as \textit{contact labels}, which are a form of weak label \cite{paul2020domain}. A contact label consists of the regions of the hand that are in contact with a surface and the level of applied force.

We collect a dataset, \datasetName{}, which captures 51 participants applying pressure to surfaces with their hands. The dataset contains \textit{fully labeled data}, which captures participants interacting with a pressure sensor. However, the sensor reduces the diversity of data that can be collected. We also capture \textit{weakly labeled data}, which is captured without a pressure sensor but contains greater diversity.

Training our network, \networkName{}, on RGB images paired with contact labels results in higher performance on diverse surfaces, outperforming prior work and generalizing to surfaces that are not represented in the fully labeled training data.

Finally, we demonstrate an application of \networkName{} to mixed reality. Visual pressure estimation allows using everyday surfaces as touch-sensitive user interfaces. We demonstrate a variety of interfaces, including a touch-sensitive keyboard that allows users to quickly type by touching a table surface. Participants type faster and prefer our keyboard when compared to a commercially released pose-based keyboard included with the Meta Quest 2 headset.

In summary, we make the following contributions:

\begin{itemize}
    \item We present \networkName{}, a deep model for pressure estimation that leverages contact labels to learn from data with and without ground truth pressure labels.
    \item We collect \datasetName{}, a dataset of RGB images with 51 participants interacting with 100+ surfaces. 
    \item We demonstrate the utility of \networkName{} with applications to mixed reality.
    \item We release our models, data, and code.
\end{itemize}


\section{Related Work}

\newcommand{\y}{\textbf{\checkmark}}
\newcommand{\n}{$\mathbf{\times}$}
\begin{table*}
\centering
\resizebox{15cm}{!}{%
\begin{tabular}{c|c|c|c|c|c|c|c|c|c}
    \textbf{} & \textbf{} & & \textbf{Partici-} & \textbf{Objects /} & \textbf{} & \textbf{} & & \textbf{Whole-} & \textbf{Natural} \\
    \textbf{Dataset} & \textbf{Modality} & \textbf{Frames} & \textbf{pants} & \textbf{Surfaces} & \textbf{Contact} & \textbf{Pressure} & \textbf{Pose} & \textbf{Hand} & \textbf{Objects} \\\hline

    OakInk \cite{yang2022oakink} & RGBD & 230k & 12 & 100 & Inferred from pose & \n  & \y & \y & \y \\\hline
    DexYCB \cite{chao2021dexycb} & RGBD & 582k & 10 & 20 & Inferred from pose & \n & \y & \y & \y \\\hline
    HO-3D \cite{hampali2020honnotate} & RGBD & 78k & 10 & 10 & Inferred from pose & \n & \y & \y & \y \\\hline
    GRAB \cite{GRAB:2020} & Pose & 1.6M & 10 & 51 & Inferred from pose & \n & \y & \y & \n \\\hline
    ContactPose \cite{brahmbhatt2020contactpose} & RGBD & 3.0M & 50 & 25 & Thermal imprint & \n & \y & \y & \n \\\hline
    PressureVisionDB \cite{grady2022pressurevision} & RGB & 3.0M & 36 & 2 & Pressure sensor & \y & \n & \y & \n \\\hline\hline
    \datasetName{} (ours) & RGB & 2.9M & 51 & 106 & Pressure sensor & \y & \n & \n & \y \\\hline
\end{tabular}
}
\vspace{-1mm}
\caption{Several hand/object datasets infer contact from pose. However, this requires accurate pose tracking and hand/object models, limiting the quality of the inferred contact. ContactPose \cite{brahmbhatt2020contactpose} captures the heat left by hands grasping objects. \datasetName{} features ground truth pressure measurements, a large number of participants, and interaction with natural objects, but only captures fingertip contact.}
\label{tab:dataset_compare}
\end{table*}



\paragraph{Physical Sensors for Pressure Sensing:} Sensors to measure pressure may be mounted to human hands.
Glove-based sensors have been developed by researchers \cite{buscher2015flexible, DBLP:journals/nature/SundaramKLZ0M19} and are commercially available \cite{pps_tactileglove, tekscan}. However, sensors mounted to the hand are expensive, interfere with tactile perception, and impact manual dexterity. Further, gloves occlude the surface of the hand, which interferes with data collection for visual models intended for bare hands. 

Various types of pressure sensors have been developed which can be mounted on objects, including capacitive sensors \cite{pose_from_cap_touchscreen, ahuja2021touchpose, guo2015capauth}, force-sensitive resistors \cite{senselmorph, brahmbhatt2020grasp, pham2017hand}, flexible sensors \cite{bhirangi2021reskin, tekscan, kim2011capacitive}, and fabric-based sensors \cite{luo2021learning}. However, even flexible pressure sensors have difficulty in conforming to the complex geometry of everyday objects. Mounting pressure sensors to objects also fundamentally alters their visual appearance and mechanical properties, reducing the diversity of data that can be captured.

\paragraph{Visual Hand Pressure Estimation:} The net forces applied by a hand to a known object can be estimated by observing the object's pose over time and calculating the forces that would result in this trajectory \cite{ehsani2020use,li2019motionandforce, pham2015forcesensing,pham2017hand,rogez2015understanding}. These methods can infer contact that is occluded or out-of-view, but they fail for static objects like tabletops. Contact estimates based on mesh geometry are also sensitive to precision since contact depends on millimeter scale displacements \cite{contactopt}. 

A number of approaches have demonstrated that visual cues can be used to estimate hand pressure, including fingertip color changes~\cite{fingertip_color,mascaro2001photoplethysmograph,mascaro2004}, soft tissue deformation~\cite{hwang2017inferring}, and cast shadows~\cite{hu2002visual,hu2000visual,cast_shadows_psychology}. In contrast to this prior work, our method uses an external camera to view the whole hand from a distance and deep learning to take advantage of multiple types of cues. Our method builds on PressureVision~\cite{grady2022pressurevision}.

\paragraph{Deep Learning with Weak Labels:} In cases where full labels are not available, approaches have been developed to still use partially or \textit{weakly} labeled data. Prior work has used semantic segmentation as a motivating task, where generating per-pixel labels requires significant time from human annotators \cite{cordts2016cityscapes}. Techniques have been developed to leverage faster annotations, including image-level labels \cite{ahn2018learning, chang2020weakly, kolesnikov2016seed} and point labels \cite{bearman2016s}. Most similar to our paper is work that leverages image-level labels and an adversarial loss to transfer segmentation models to new domains \cite{paul2020domain}. In contrast to weak labels applied by human annotators after data collection, our method prompts human behavior while data is being collected.



\paragraph{Hands in Mixed Reality:} Modern mixed reality devices increasingly rely on hand tracking as an input modality. Commercially available devices use monochrome cameras \cite{han2022umetrack, Han2020MEgATrackME} and depth sensors \cite{guo2022hololens} to estimate 3D hand pose.

In order to sense contact with the environment, a variety of hand-mounted physical sensors have been proposed, including IMUs \cite{shi2020ready, meier2021tapid}, electrical current injection \cite{zhang2019actitouch, kienzle2021electroring}, and acoustic sensing \cite{gong2020acustico}. Most similar to our work is research in sensing contact between fingers and flat surfaces from depth cameras \cite{mrtouch, shen2021farout, goussies2023learning}. However, depth cameras add cost, draw high power, and may not work on reflective surfaces or in brightly illuminated scenes. Time-of-flight cameras may have error on materials such as human skin \cite{he2017depth}. In comparison, our work senses pressure from only RGB cameras, which are non-invasive and low-cost.

\begin{figure*}
  \centering
  \includegraphics[width=1.0\linewidth]{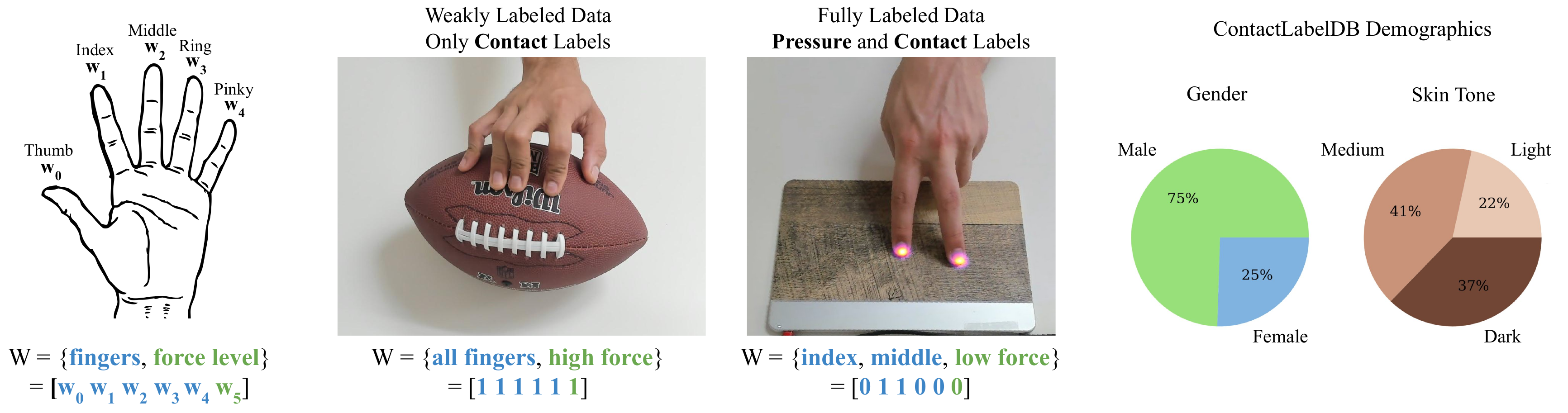}
  \caption{We represent the contact labels as a six-dimensional vector. The first five elements are binary values indicating which fingers are prompted to be in contact, and the last element indicates the prompted force level. Fully labeled data has \textit{both} pressure and contact labels, while weakly labeled data has \textit{only} contact labels. Participants with a range of genders and skin tones were recruited for our study.}
  \label{fig:weak_explanation}
\end{figure*}

\section{Data Collection}



This section describes the capture of \datasetName{}, a dataset of 51 participants making contact with diverse surfaces and objects. 






\subsection{Contact Labels}

During all data capture sequences, the participant is prompted to make contact with a surface using a specific combination of fingertips and to achieve a target force level, for example: ``press ring finger at a low force". The participant performs the requested action while data is collected. This prompt has a one-to-one correspondence with a contact label (Figure \ref{fig:weak_explanation}).

As shown in Figure \ref{fig:weak_explanation}, a contact label $W$ is represented as a vector with 6 elements. The first 5 elements indicate the presence or lack of contact at each of the 5 fingertips. The sixth element indicates if the participant was prompted to exert a low, high, or unspecified force with the fingers in contact.

We represent a contact label $W \in \mathbb{Z}^6$ as follows:
\begin{align*}
w_i|_{0 \leq i \leq 4} \in \{0, 1\} &\equiv \{\textit{no contact, contact}\}\\
w_i|_{i = 5} \in \{-1, 0, 1\} &\equiv \{\textit{unspecified, low, high force}\}
\end{align*}

For all data collection procedures, we prompted participants to press one of eight combinations of fingertips onto a surface. For each combination, we prompted the participant to apply a low force, a high force, or to slide with unspecified force. Additionally, we prompted participants to make ``no contact" by hovering the specified fingertips just above the surface.

While PressureVisionDB \cite{grady2022pressurevision} captured contact with the entire hand including the palm, in this work, we only capture fingertip contact. This decision was made since prompting contact with other parts of the hand is complex, and fingertips are sufficient for many downstream tasks. During pilot studies, we also found that participants apply similar pressures between fingertips unless explicitly prompted not to.

\subsection{Collection Method} \label{sec:data_collection_technique}


We collect two types of data: \textit{fully labeled} data and \textit{weakly labeled} data.
For both types of sequences, we collect contact labels by prompting the subject with a specific instruction. For the fully labeled sequences, we additionally collect ground truth pressure labels using a high-resolution pressure sensing array~\cite{senselmorph} (Figure \ref{fig:weak_explanation}).

When data is collected with a ground truth pressure sensor, participants press and release their hand multiple times on the surface to capture the onset and termination of contact. Frames with no pressure detected are assigned a ``no-contact" label. When data is collected without a pressure sensor, the participant maintains contact throughout the duration of the recording.

\subsection{Data Splits} \label{sec:data_splits}
Our training data comes from 37 participants, while our testing data comes from 14 participants who are not present in the training data. Since weakly labeled data collection does not require the pressure sensor, a much greater diversity of data can be collected. We collect a total of 2.9M frames: 0.5M fully labeled frames and 2.4M weakly labeled frames.

Our training data consists of a \textbf{fully labeled training set} where participants interacted with solid-colored overlays on a pressure sensor. We also collect a \textbf{weakly labeled training set} where participants interacted with diverse surfaces and natural objects instead of a pressure sensor.

We desire to evaluate our approach on interactions with the natural world. However, we face the same problem as during training data collection: it is difficult to collect ground truth pressure in diverse environments. We collect a \textbf{fully labeled testing set} where participants interact with textured overlays not seen in the training set on a pressure sensor. We also collect a \textbf{weakly labeled testing set} where participants interact with diverse surfaces and natural objects, many of which were not present in the training set.

Images were captured with multiple consumer-grade webcams at 30 FPS and 1080p resolution. We conducted data collection in 20 environments with different lighting conditions. We used a Sensel Morph \cite{senselmorph} pressure sensing array. 




\begin{figure*}
  \centering
    \includegraphics[width=0.85\linewidth]{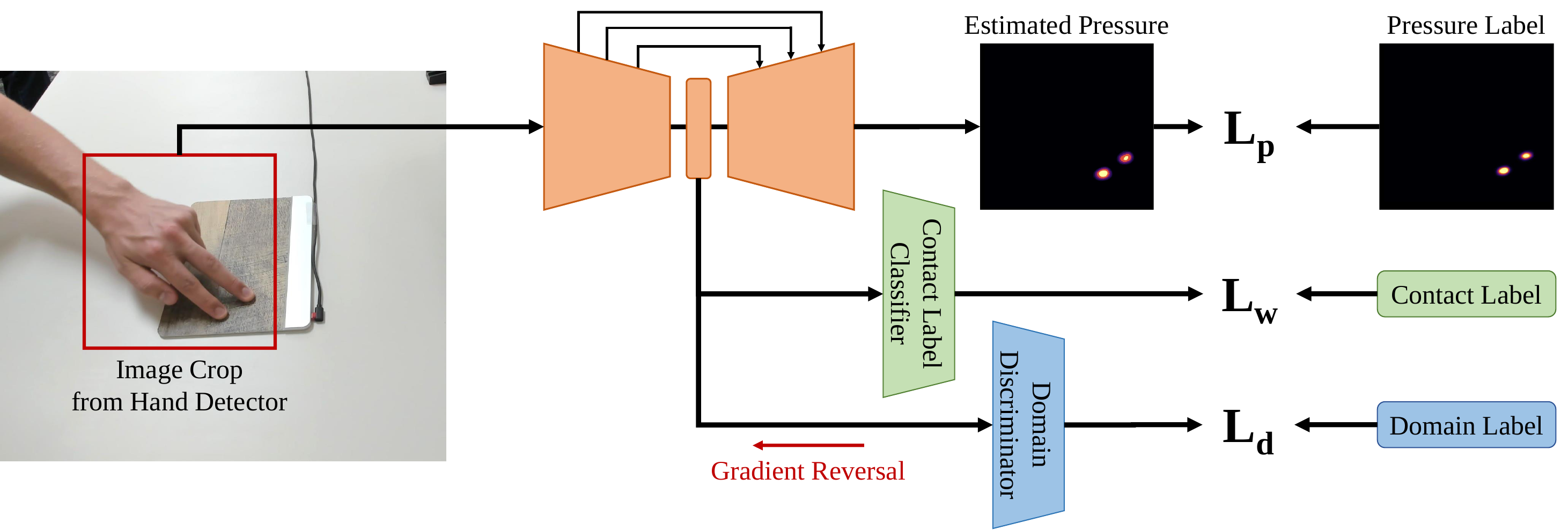}
  \caption{\networkName{} architecture. First, hand crops are generated using the bounding boxes estimated by an off-the-shelf hand detector. The crops are passed into an encoder-decoder network to estimate pressure for each pixel in the input image. Two classification heads are attached to the bottleneck of the network; one is trained to estimate the contact label, and the other uses an adversarial loss to reduce the shift between fully labeled and weakly labeled domains.}
  \label{fig:network_diagram}
\end{figure*}

\subsection{Ethics}
Approval to conduct this study was obtained from an Institutional Review Board (IRB). We recruited a diverse set of 51 participants (Figure \ref{fig:weak_explanation}). All participants gave informed consent and were compensated for their time. We measured skin tone with a Pantone X-Rite RM200 spectrocolorimeter, and participants self-reported gender. 






\section{Network Architecture}


We create a network, \networkName{}, (Figure \ref{fig:network_diagram}) to take a single RGB image, $I$, as input and output a pressure image, $\hat{P}=f(I)$. For fully labeled data, each RGB image is paired with a ground-truth pressure image obtained by projecting the output of a pressure-sensing array into the image using a homography transform. The output pressure $\hat{P}$ is in \textit{image space}, such that the input and output images are the same shape and can be superimposed (Figure \ref{fig:collage_full}).

\subsection{Pressure Estimation}

To estimate pressure, \networkName{} uses a binned representation and performs classification across bins. The pressure range is split into $N_B=9$ logarithmically spaced bins divided across the pressure range, including one zero bin. Pressure estimation uses a \textit{structure-aware cross-entropy} loss $L_p$ \cite{su2015render, massa2016crafting}. Unlike regular cross-entropy, the structure-aware loss penalizes large errors more than small errors. For each pressure pixel over the image $x,y$, the loss is computed over all bin indices $b\in B$ using the ground truth index $k_b$ and the estimated probability for each bin $\rho_{x,y}(b)$.
\begin{align}
    L_p = -\sum_{x,y}\sum_{b}e^{-|b-k_b|}log(\rho_{x,y}(b))
\end{align}
$L_p$ is only computed when fully labeled data is available.

\subsection{Contact Label Estimation}

In addition to estimating a pressure image, \networkName{} performs the auxiliary task of estimating the contact label $\hat{W}$. The contact label classifier predicts $\hat{W}$ given the features $F$ at the network bottleneck (Figure \ref{fig:network_diagram}). The addition of the contact label classifier ensures that the features generated by the encoder are discriminative to the set of fingers in contact and the force level. The classifier pools features and uses a 2-layer MLP to estimate the contact label collected in Section \ref{sec:data_collection_technique}. This classifier is trained with a binary cross-entropy loss $L_w$. When the amount of force is not specified in the ground truth contact label, this portion of the loss is masked out.


\subsection{Adversarial Domain Adaptation}
Following prior work in domain adaptation \cite{ganin2016dann}, we apply an additional feature alignment loss using a domain discriminator $D$. This loss is \textit{unsupervised}, as it does not leverage contact label information. This loss attempts to minimize the distance between the distributions of features generated from two domains. The discriminator estimates if the image is from the fully labeled or weakly labeled domain, and when backpropagating, gradients are reversed upstream of the domain discriminator \cite{ganin2016dann}. For image features from the fully labeled domain $F_f$ and weakly labeled domain $F_w$, the domain loss function $L_d$ is:
\begin{align}
    L_d = -log(D(F_f)) -log(1-D(F_w))
\end{align}

\subsection{Training Details}
As hands often only take up a small part of the image, \networkName{} operates on crops of the hand. We use MediaPipe \cite{mediapipe} to produce bounding boxes which are used to generate hand crops. Hand crops are resized to 448x448 pixels before being sent to \networkName{}.

\networkName{} uses an SE-ResNeXt-50 encoder \cite{resnet,squeeze-excitation,resnext} and an FPN decoder \cite{fpn,segmentation_models_pytorch}, and is trained end-to-end using the following loss function:
\begin{align}
    L = L_p + \lambda_1 L_w + \lambda_2 L_d
\end{align}

\begin{figure*}
  \centering
  \includegraphics[width=0.75\linewidth]{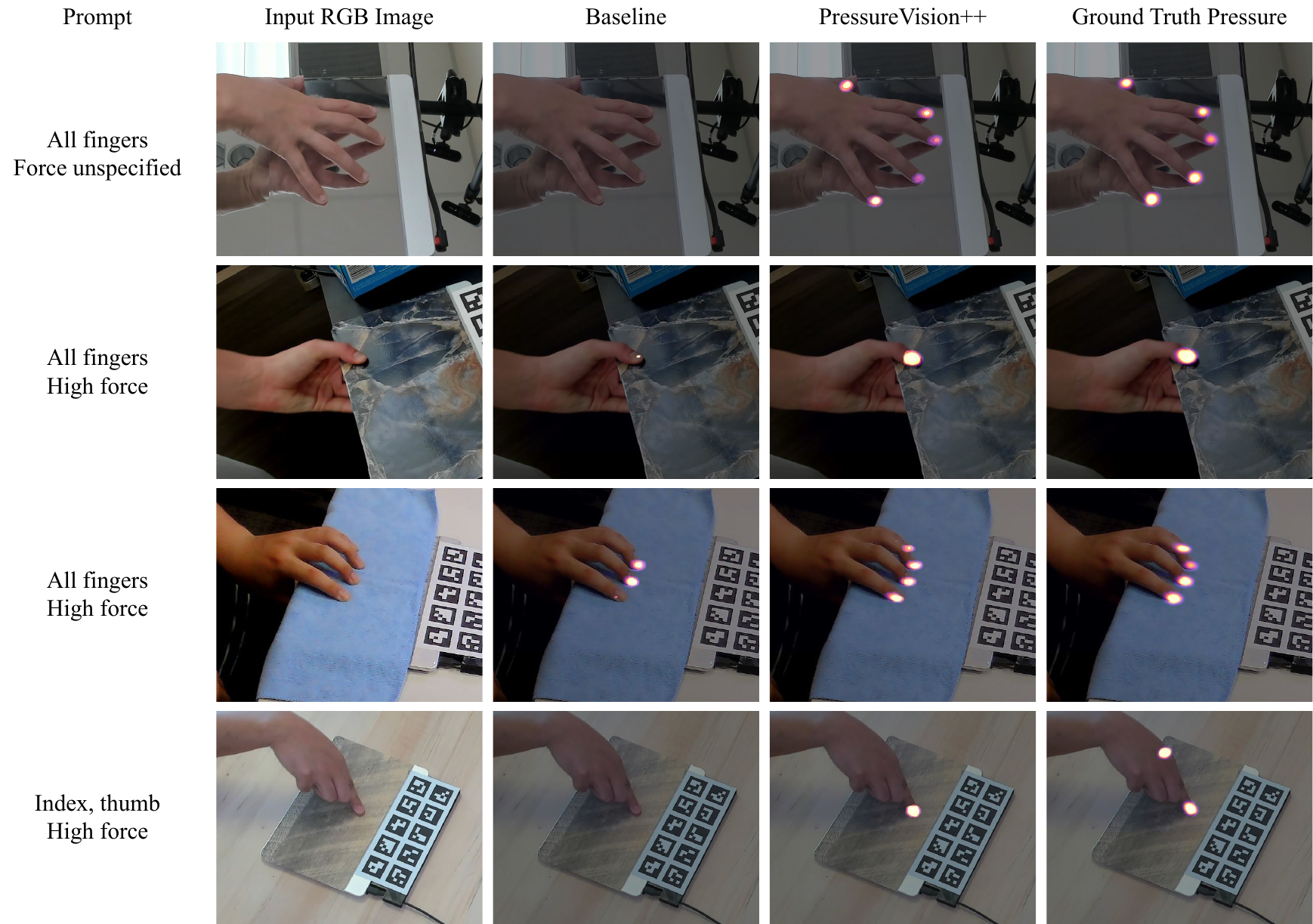}
  \caption{Results on the fully labeled test set. The baseline column is \networkName{} trained without either the domain loss or contact label loss. The bottom row shows a common failure mode where pressure is not estimated for occluded parts of the hand.}
  \label{fig:collage_full}
\end{figure*}

\begin{table*}
\centering
\begin{tabular}{c|c|c|c||c}
      & \multicolumn{3}{c||}{\textbf{Fully Labeled Test Set}} & \textbf{Weakly Labeled Test Set} \\\hline
    \textbf{Method} & \textbf{Contact Acc.} & \textbf{Contact IoU} & \textbf{Volumetric IoU} & \textbf{Contact Acc.} \\\hline
    Zero Guesser & 53.4\% & 0.0\% & 0.0\% & 24.9\% \\\hline
    Human Annotator & 78.4\% & - & - & \textbf{80.5\%} \\\hline\hline
    PressureVision \cite{grady2022pressurevision} & 72.7\% & 15.2\% & 11.3\% & 53.5\% \\\hline
    \networkName{} (ours) & \textbf{89.3\%} & \textbf{41.9\%} & \textbf{27.5\%} & \textbf{80.5\%} \\\hline
\end{tabular}
\vspace{1mm}
\caption{Performance compared to a PressureVision baseline~\cite{grady2022pressurevision} and human annotators.}
\label{tab:main_results}
\end{table*}
\section{Evaluation}




We consider two types of evaluations: \textit{contact} and \textit{pressure} evaluations, following prior work \cite{grady2022pressurevision}. Contact is a binary quantity indicating if the hand and object are touching, while pressure is a scalar indicating the magnitude of force. A binary contact image $\hat{C}$ is generated by thresholding each pressure pixel in $\hat{P}$ at $P_{th}=1~kPa$.

\begin{itemize}
    \item \textbf{Contact Accuracy:} The estimated contact image $\hat{C}$ is used to determine if \textit{any} contact is estimated. Accuracy is calculated by counting the percentage of video frames for which $\hat{C}$ corresponds with the contact label.
    \item \textbf{Contact IoU:} Intersection-over-union (IoU) is computed between the ground truth contact image $C$ and estimated contact image $\hat{C}$. 
    \item \textbf{Volumetric IoU:} An extension of Contact IoU that considers the \textit{magnitude} of pressure. 2D pressure images are viewed as 3D pressure volumes, where the height of the volume is equal to the magnitude of pressure at that pixel. Intersection-over-union is computed using these volumes.
\end{itemize}

For the same reasons that collecting fully labeled training data on diverse surfaces is difficult, collecting fully labeled testing data also presents challenges. We evaluate both the \textit{fully labeled} and \textit{weakly labeled} test sets. However, due to the lack of pressure measurements in the weakly labeled test set, only contact accuracy is computed. For more details and evaluations, refer to the supplementary material.


\begin{figure*}
  \centering
  \includegraphics[width=0.95\linewidth]{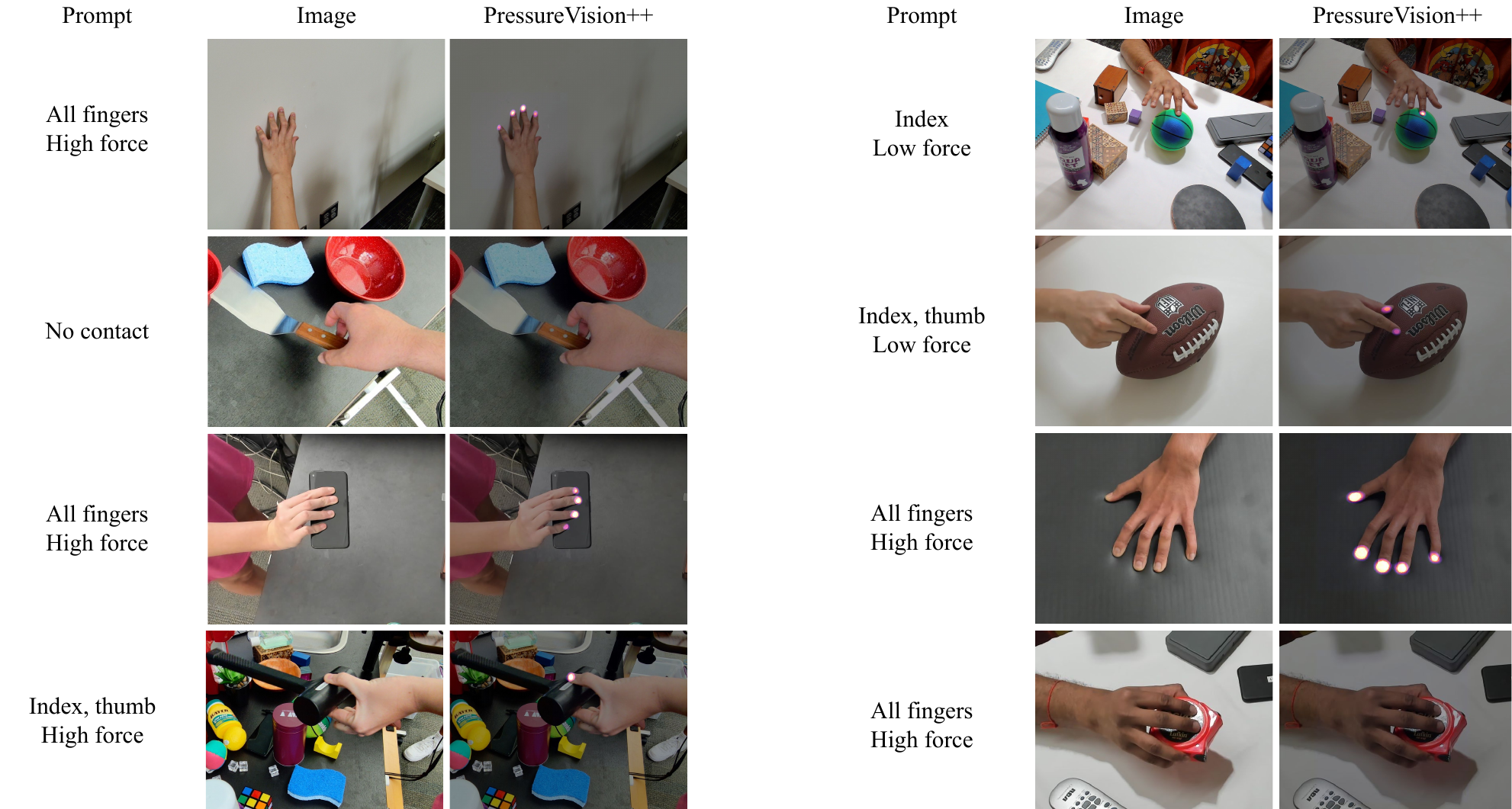}
  \caption{Results on the surfaces in the weakly labeled test set, \textit{none} of which are included in the fully labeled training set. \networkName{} produces qualitatively accurate results on highly textured, curved, and compliant surfaces. All images except the top row are zoomed in to show detail. The bottom right shows a failure case where pressure is underestimated on an object.
  }
  \label{fig:weak_collage}
\end{figure*}

\subsection{Performance Compared to Baselines} 

We compare our method against three baselines as shown in Table \ref{tab:main_results}.
\begin{itemize}
\item \textbf{Zero Guesser:} The zero guesser outputs a zero-pressure image and provides a reference for Contact Accuracy due to the large number of frames with no contact.
\item \textbf{Human Annotator:} Annotators from Amazon Mechanical Turk perform a binary classification on 10,000 images to determine if any part of the hand is touching the surface.

\item \textbf{PressureVision:} The network from \cite{grady2022pressurevision} is retrained on our fully labeled data. This method does not use contact labels.
\end{itemize}

\networkName{} significantly outperforms prior work, improving on all metrics. Examples from the fully labeled test set are shown in Figure~\ref{fig:collage_full}, and examples from the weakly labeled test set are shown in Figure \ref{fig:weak_collage}. \networkName{} estimates fingertip pressure on diverse surfaces, including textured, deformable, and curved surfaces. Our method adapts to unseen surfaces in the test set by leveraging diverse weakly labeled data. We observe one common failure mode where pressure is not estimated for occluded fingertips (Figure \ref{fig:collage_full}).


\paragraph{Human-Annotated Contact:} We investigated the possibility of using non-expert human labelers recruited from Amazon Mechanical Turk to identify contact from images. In this evaluation, workers performed a binary classification on 10,000 frames to identify whether the hand is in contact or not. This is an easier task than generating contact labels or estimating pressure.

We observe that non-expert annotators had difficulty distinguishing near-contact from contact (Table \ref{tab:main_results}). When contact accuracy is computed, we observe that \networkName{} outperforms the human annotators on the fully labeled test set, and performs similarly on the weakly labeled test set. This result suggests that our approach may exceed human performance under ideal conditions, but is not as robust when tested on more diverse data.

\subsection{Ablating Weakly Labeled Data}

\label{sec:ablations}
Table \ref{tab:data_ablations} illustrates the impact of weakly labeled data on \networkName{}'s performance. With neither the domain loss nor contact label loss, the weakly labeled dataset is unused. With only the domain loss $L_d$, the weakly labeled dataset is used in an unsupervised way. Finally, the contact label loss $L_w$ leverages the contact labels collected. We find that both losses significantly contribute to performance. However, the contact label loss has the largest effect, with this alone improving volumetric IoU by +71\%. 

This large performance improvement demonstrates the value of contact labels for hand pressure estimation. Weakly labeled data is easy to collect, yet significantly increases performance on diverse surfaces.

\begin{table}
\centering
\begin{tabular}{c|c|c}
    \textbf{$L_d$} & \textbf{$L_w$} & \textbf{Volumetric IoU} \\\hline
    &  & 14.9\% \\\hline
    \y &  & 17.5\% \\\hline
    & \y & 25.5\% \\\hline
    \y & \y & \textbf{27.5\%} \\\hline
\end{tabular}
\vspace{1mm}
\caption{The domain loss $L_d$ and contact label loss $L_w$ enable training on weakly labeled data which improves performance significantly.}
\label{tab:data_ablations}
\end{table}
\section{Applications in Mixed Reality}

Modern mixed reality devices increasingly rely on hand tracking as a primary input modality. The Meta Quest and Microsoft HoloLens product lines use 3D pose estimation to allow hands to interact with mid-air interfaces. However, mid-air interfaces are fatiguing, and virtual objects do not provide tactile feedback. A study by Cheng \etal \cite{cheng2022comfortable} compared mid-air interfaces to tabletop interfaces. Participants interacting on the tabletop were more accurate and reported less exertion and improved comfort.

\methodName{} presents a natural way to extend hand tracking to detect interactions with surfaces. Our system only requires a low-cost, externally mounted RGB camera and may enable more accurate, lower-exertion input.

\begin{figure*}
  \centering
  \includegraphics[width=1.0\linewidth]{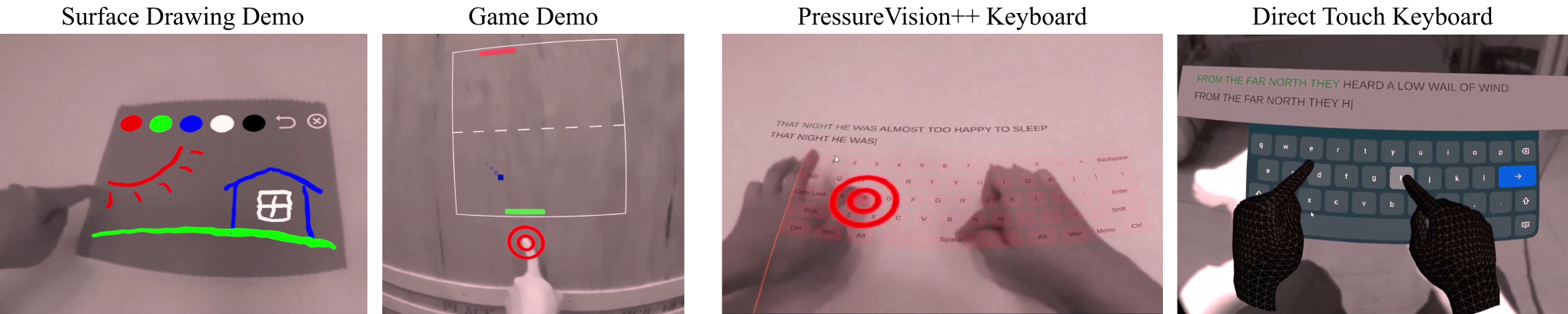}
  \vspace{-2mm}
  \caption{\methodName{} allows touch-sensitive interfaces to be placed on objects. We show drawing on the back of a notebook and playing a game on a vertical wooden surface. The \methodName{} Keyboard allows users to type by touching keys on a tabletop. This is compared to the Direct Touch Keyboard which uses 3D hand poses to allow users to press keys of a mid-air keyboard.}
  \label{fig:typing}
\end{figure*}



\subsection{Hardware Setup}

We use a Meta Quest 2 headset for our demos. The Quest operates in passthrough mode which allows users to ``see through" the headset by using the device's onboard cameras to provide a mixed reality experience. However, we do not use these onboard cameras for pressure estimation.

We mount a single RGB camera to a table (Figure 1) and run \methodName{} on a desktop computer. The pressure estimation system runs at approximately 50 FPS with an RTX 3090 GPU, which includes first running the hand detector \cite{mediapipe} and then running \networkName{} once for each detected hand. Pressure information is sent over WiFi to the headset. See the supplementary materials for more details.

\subsection{Touch-Sensitive Interfaces}
\methodName{} enables the creation of touch-sensitive user interfaces which can be attached to tabletops, walls, or objects in the environment (Figure \ref{fig:typing}). We developed a pressure-sensitive drawing application that allows users to paint on a surface simply by dragging their finger. The amount of pressure they apply controls the width of the brush stroke. Sample drawings are shown in Figures 1 and \ref{fig:typing}. We also design a game similar to Pong where the user controls the paddle position by dragging their finger and show a user interacting with this game on a vertical wooden surface.

\subsection{Touch Typing}

To demonstrate the accuracy and responsiveness of our approach, we evaluate \methodName{} on a typing task. We develop a touch-sensitive keyboard using our system which allows users to tap on a tabletop to enter text.

\textbf{\methodName{} Keyboard:} The headset projects a keyboard layout onto a tabletop. When the system detects contact, a red marker is drawn (Figure \ref{fig:typing}) to aid the user. We implement a debouncing filter that requires 2 subsequent frames to trigger the onset or termination of a keypress, reducing spurious keystrokes.

\textbf{Baseline Keyboard:} The Quest 2 headset performs 3D hand pose estimation using its four monochrome cameras \cite{han2022umetrack}. The Meta Direct Touch Keyboard is a text entry method that places a floating keyboard in front of the user and allows users to press keys using their index fingers (Figure \ref{fig:typing}). As this keyboard is built-in to the Quest 2 operating system, we use it as a baseline.

\textbf{User Study:} To evaluate the typing methods, we recruited 10 participants, none of whom participated in the collection of \datasetName{}. The two typing methods were presented in random order. Participants attempted to quickly and accurately type a prompt sentence. Typing speed was calculated in net words per minute \cite{salthouse1984effects}. After completing the typing test, participants gave open-ended responses comparing the two systems. They also selected which keyboard they preferred.

\begin{table}
\centering
\begin{tabular}{c|c|c}
    \textbf{Typing Method} & \textbf{Net WPM} & \textbf{Preferred} \\\hline
    Direct Touch & 14.4 & 1/10 \\\hline
    \methodName{} (ours) & \textbf{25.8} & \textbf{9/10} \\\hline
\end{tabular}
\caption{Participants type faster with the \methodName{} Keyboard as compared to the Direct Touch Keyboard. After trying both, 9 of 10 participants preferred our method.}
\label{tab:typing_results}
\end{table}

\textbf{Results:} We found that participants typed 78\% faster and 9 of 10 participants favored the \methodName{} Keyboard over the Direct Touch Keyboard (Table \ref{tab:typing_results}). In their open-ended responses, 7 participants mentioned that they found the tactile feedback of interacting with a real surface helpful, and 5 mentioned that they found the \methodName{} keyboard less tiring to use.

Overall, these results suggest that surface interactions enabled by \methodName{} have advantages over the mid-air interfaces enabled by pose estimators. 

\textbf{Limitations:} We observe that during 5-finger typing, fingertips are often occluded and users make hand poses that are not represented in our dataset. As a result, we instructed participants to only type with their index fingers, which are more reliably detected. Additionally, compared to the Direct Touch Keyboard, our approach is not limited to the monochrome egocentric cameras and the mobile processor that the headset uses for hand pose estimation. However, these limitations may be overcome in future work. 

\section{Conclusion}

Training deep models to visually estimate the pressure applied by fingertips relies on ground-truth pressure measurements that are difficult to obtain. We presented \methodName{} which uses more easily obtained contact labels collected by prompting participants to achieve specific types of contact. Leveraging this weakly supervised data improves pressure estimation on diverse surfaces and outperforms prior methods. \methodName{} additionally enables interactions with natural surfaces in mixed reality.


{\small
\bibliographystyle{ieee_fullname}
\bibliography{egbib}
}



\vskip .375in

\twocolumn[{%
\renewcommand\twocolumn[1][]{#1}%

\begin{center}
  {{\Large \bf Supplementary - PressureVision++: Estimating Fingertip Pressure\\from Diverse RGB Images \par}}
  {\vspace*{24pt}}
  {
  \large
  \lineskip .5em
  \begin{tabular}[t]{c}
      Patrick Grady\textsuperscript{1}, Jeremy A. Collins\textsuperscript{1}, Chengcheng Tang\textsuperscript{2}, Christopher D. Twigg\textsuperscript{2},\\Kunal Aneja\textsuperscript{1}, James Hays\textsuperscript{1}, Charles C. Kemp\textsuperscript{1}\\
\\
\textsuperscript{1}Georgia Institute of Technology,~\textsuperscript{2}Meta Reality Labs\\
  \end{tabular}
  \par
  }
  \vskip .5em
  \vspace*{12pt}
\end{center}

}]



\section{Introduction to Supplementary}

This supplementary document provides additional details and results that were not included in the main paper. Section \ref{sec:data_collection_details} provides additional details surrounding the data collection, Section \ref{sec:network_arch_details} provides additional details surrounding the network architecture and training, Section \ref{sec:ar} provides additional details about the mixed reality applications. Additional results are shown in Figures \ref{fig:supp_collage_full_1}, \ref{fig:supp_collage_full_2}, \ref{fig:supp_collage_weak_1}, and \ref{fig:supp_collage_weak_2}.

\section{Data Collection Details}
\label{sec:data_collection_details}


\subsection{Actions}

Participants were prompted with actions from a list of prompts, shown in Table \ref{tab:action_prompts}. Most of the actions were repeated across four force levels: low force, high force, slide (force unspecified), and no contact. Due to the highly varying frictional properties of each surface, we did not prompt a force level during the slide prompt. Not all participants completed all actions.

\subsection{Data Collection Hardware}

To record pressure, a Sensel Morph \cite{senselmorph} pressure sensor was used. This sensor records a $105\times 185$ pressure image. To vary the sensor's appearance, various commercially available adhesive vinyl coverings were applied to the sensor's active area. The location and lighting were also changed to vary exposure (and thus the amount of motion blur), hue, and saturation of the images.

Data was captured from seven consumer-grade webcams, including four Logitech Brio 4K webcams, one Dell Ultrasharp 4k webcam, one Elgato Facecam 1080p webcam, and one Lumina 4k webcam. All streams were recorded at 1080p and 30 FPS, and later down-sampled to 15 FPS due to the large size of the dataset.

Most of the data was captured under unaltered room lighting, however some was collected in a room illuminated with smart LED bulbs which randomly changed brightness, providing a greater diversity of lighting. The data collection took place in twenty different environments.

For recordings with the ground truth pressure sensor, the cameras were spatially calibrated with an ArUco board \cite{garrido2014automatic}. The cameras were temporally aligned with pressure sensor readings with a specialized tool. When pressed against the pressure sensor, the tool would illuminate, allowing the pressure readings and camera frames to be aligned.


\subsection{Dataset Statistics}

\begin{table}
\centering
\begin{tabular}{c|c}
    Participants & 51 \\\hline
    Cameras & 7 \\\hline
    Objects & 106 \\\hline
    Locations & 20 \\\hline
    Resolution & 1920x1080 \\\hline
    Framerate & 15 FPS \\\hline
    Total Frames & 2.9M \\\hline
    Full Train Frames & 182k \\\hline
    Weak Train Frames & 1805k \\\hline
    Full Val Frames & 21k \\\hline
    Weak Val Frames & 72k \\\hline
    Full Test Frames & 305k \\\hline
    Weak Test Frames & 509k \\\hline
    Mean force, high force prompt & 19.6N \\\hline
    Mean force, low force prompt & 3.6N \\\hline
\end{tabular}
\caption{\datasetName{} Statistics.}
\label{tab:dataset_results}
\end{table}

\begin{figure}
  \centering
  \includegraphics[width=0.99\linewidth]{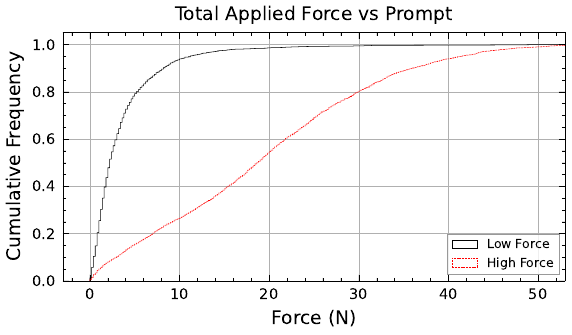}
  \caption{Cumulative distribution of total applied force versus prompt. During data collection, participants apply higher forces when prompted to apply ``high force" as opposed to ``low force".}
  \label{fig:force_distribution}
\end{figure}

We show additional information about the dataset in Table \ref{tab:dataset_results}.

During data collection, participants were prompted to apply high and low forces. Figure \ref{fig:force_distribution} shows the distribution of total applied forces as measured in the fully labeled dataset. Pressure data is integrated over the sensor area to calculate total force. This plot includes data from all sequences, meaning that the data is representative of one-finger contact as well as five-finger contact. Generally, we find a consistent difference between the two classes, and participants apply higher forces when prompted in the ``high force" case.

\section{Network and Evaluation Details}
\label{sec:network_arch_details}

\subsection{Training Details}

\networkName{} takes cropped images of the hand as input. An off-the-shelf hand detector, MediaPipe \cite{mediapipe}, generates hand bounding boxes on all images. The bounding box is used to generate a crop of the hand for \networkName{}. We discarded frames where the hand detector did not find a hand. When the participant interacted with reflective surfaces, the hand detector frequently detected both the hand and its reflection. In cases when two hands were detected, the hand towards the top of the image was chosen.

\networkName{} was implemented in PyTorch \cite{pytorch} and used network implementations from the \textit{segmentation-models-pytorch} project \cite{segmentation_models_pytorch}. \networkName{} was trained with batches of 28 images: 14 fully-labeled images and 14 weakly-labeled images. Batch size was set to fully utilize the memory of an RTX 3090 GPU. The network was optimized with the Adam optimizer \cite{adam} for 300k iterations. The learning rate was 0.001 for the first 100k iterations, and 0.0001 thereafter. Training data is augmented with horizontal flips, color jitter, random rotations, scaling, and translations.

The complete loss function is: 
\begin{align}
    L = L_p + \lambda_1 L_w + \lambda_2 L_d
\end{align}

We choose $\lambda_1=0.01$ and $\lambda_2=0.001$.

\subsection{Cross-Dataset Generalization}

In table \ref{tab:cross_dataset_results}, we show cross-dataset generalization results when \networkName{} is tested and trained on PressureVisionDB \cite{grady2022pressurevision} and \datasetName{}. Although \datasetName{} contains more diversity in terms of objects, it appears that the model trained on \datasetName{} and tested on PressureVisionDB performs worse than the model trained on PressureVisionDB and tested on \datasetName{}.

We hypothesize that this is because PressureVisionDB was captured in very harsh, artificial lighting conditions. These extreme lighting conditions are not captured in \datasetName{}, which instead captures normal indoor lighting environments. We believe that models trained on \datasetName{} generalize poorly to the extreme lighting captured in PressureVisionDB. During real-world testing, we find models trained on \datasetName{} generalize much better to real-world scenarios.

\begin{table}
\centering
\begin{tabular}{c|c|c}
    \diagbox{Train}{Test} & \textbf{PV-DB} & \textbf{CL-DB} \\\hline
    PV-DB \cite{grady2022pressurevision} & 41.3\% & 9.2\% \\\hline
    CL-DB (ours) &  2.3\% & 27.5\%  \\\hline
\end{tabular}
\caption{Cross-dataset results comparing PressureVisionDB (PV-DB) to \datasetName{} (CL-DB).}
\label{tab:cross_dataset_results}
\end{table}



\subsection{Accuracy of Estimated Contact Labels}

\networkName{} produces two outputs for every input image: a pressure image and a contact label. The main paper analyzes the accuracy of the estimated pressure image, and this section evaluates the accuracy of the estimated contact label.

We compare the performance of the pressure estimate to the contact label estimate. We report the following metrics, which are computed over both the fully labeled and weakly labeled test sets:

\begin{itemize}
    \item Contact Accuracy (pressure image) uses the estimated pressure image to determine if any contact is present across the entire image. This is compared to the ground truth contact. This is the same metric reported in Section 5 of the main paper.
    \item Contact Accuracy (contact label) uses the estimated contact label to determine if any contact is present across 5 fingers. This is compared to the ground truth contact.
\end{itemize}

\begin{table}[h]
\centering
\begin{tabular}{c|c}\hline
    Contact Accuracy (pressure image) & 83.7\% \\\hline
    Contact Accuracy (contact label) & 86.1\% \\\hline
\end{tabular}
\caption{Contact Accuracy compared between pressure estimates and contact label estimates.}
\label{tab:contact_accuracy}
\end{table}

We find that the pressure-based contact accuracy and contact-label-based contact accuracy perform similarly, with the contact-labeled-based estimate performing slightly better. 

\begin{table}
\centering
\begin{tabular}{c|c}
    \textbf{Contact Label Segment} & \textbf{Accuracy} \\\hline
    Thumb & 89.6\% \\\hline
    Index & 87.8\% \\\hline
    Middle & 90.8\% \\\hline
    Ring & 92.4\% \\\hline
    Pinky & 92.5\% \\\hline
    Force & 77.5\% \\\hline
\end{tabular}
\caption{Per-finger and force accuracy.}
\label{tab:contact_label}
\end{table}

We report per-finger contact label accuracy in Table \ref{tab:contact_label}. Force accuracy uses the estimated contact label to determine if the hand applies a high or low force. This is compared to the ground truth force level as prompted to the participant. The force accuracy is generally lower than the other segments of the contact label, suggesting that estimating the quantity of force is a more difficult task than the binary presence of contact.

\section{Applications in Mixed Reality}
\label{sec:ar}

\subsection{Surface Interactions}

\begin{figure}
  \centering
  \includegraphics[width=0.7\linewidth]{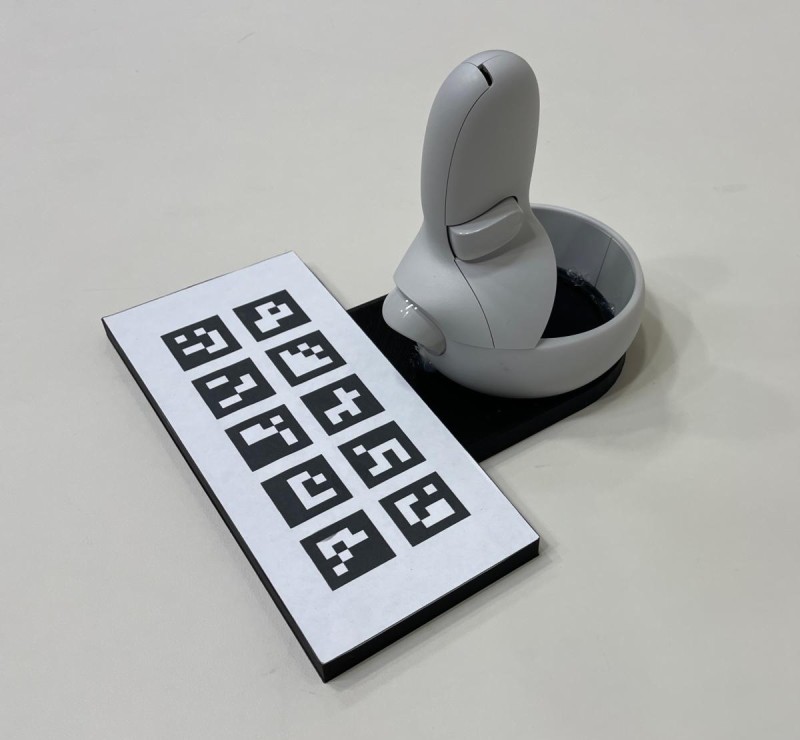}
  \caption{Calibration tool used to align coordinate frames between the RGB camera and the Meta Quest 2 headset. The controller is rigidly connected to the ArUco board.}
  \label{fig:calibration_totem}
\end{figure}

\begin{figure*}
  \centering
  \includegraphics[width=0.8\linewidth]{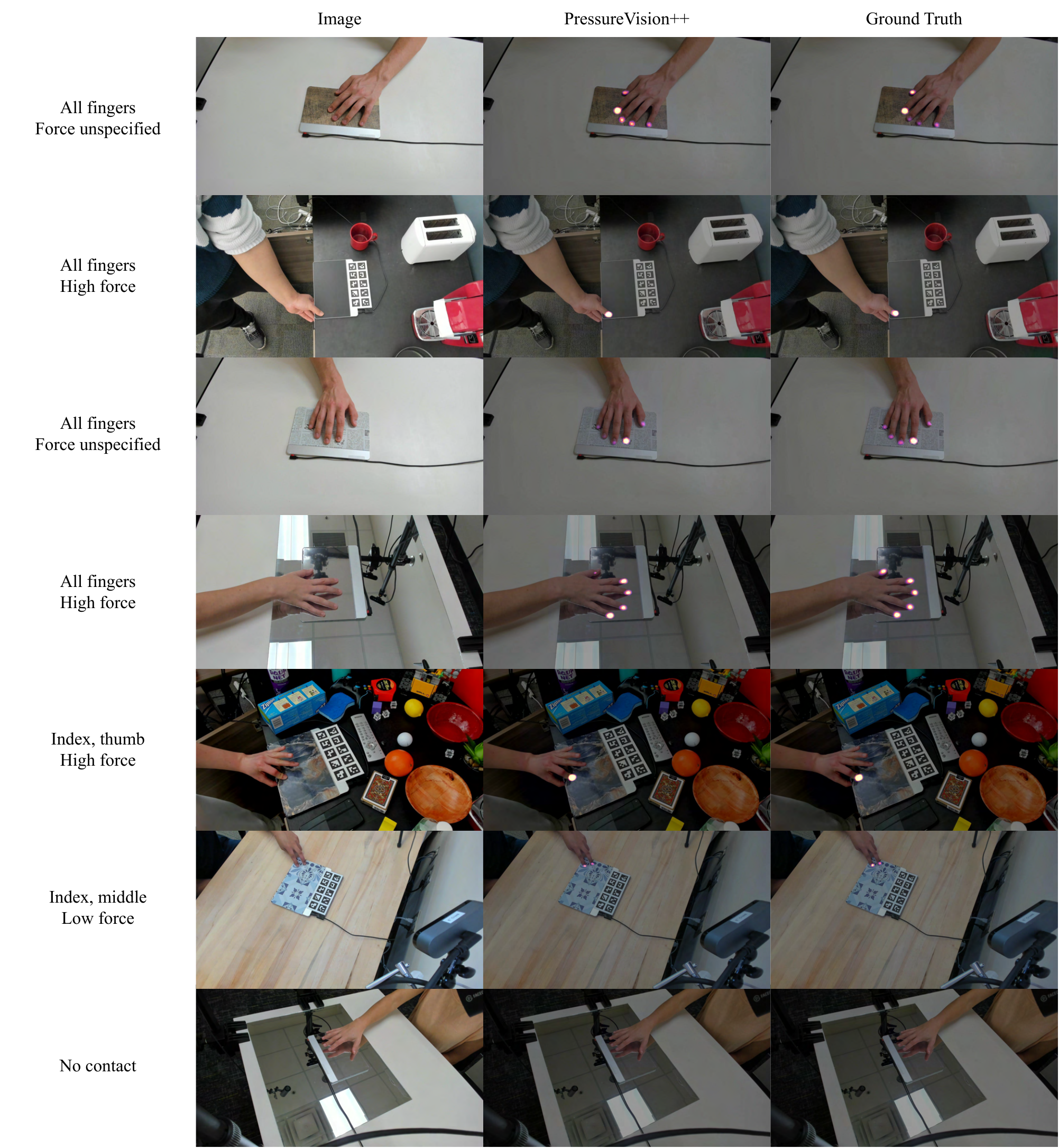}
  \caption{Results from the fully labeled test set where ground truth pressure is measured by a pressure sensor. Testing participants are held out from the training sets.}
  \label{fig:supp_collage_full_1}
\end{figure*}

\begin{figure*}
  \centering
  \includegraphics[width=0.8\linewidth]{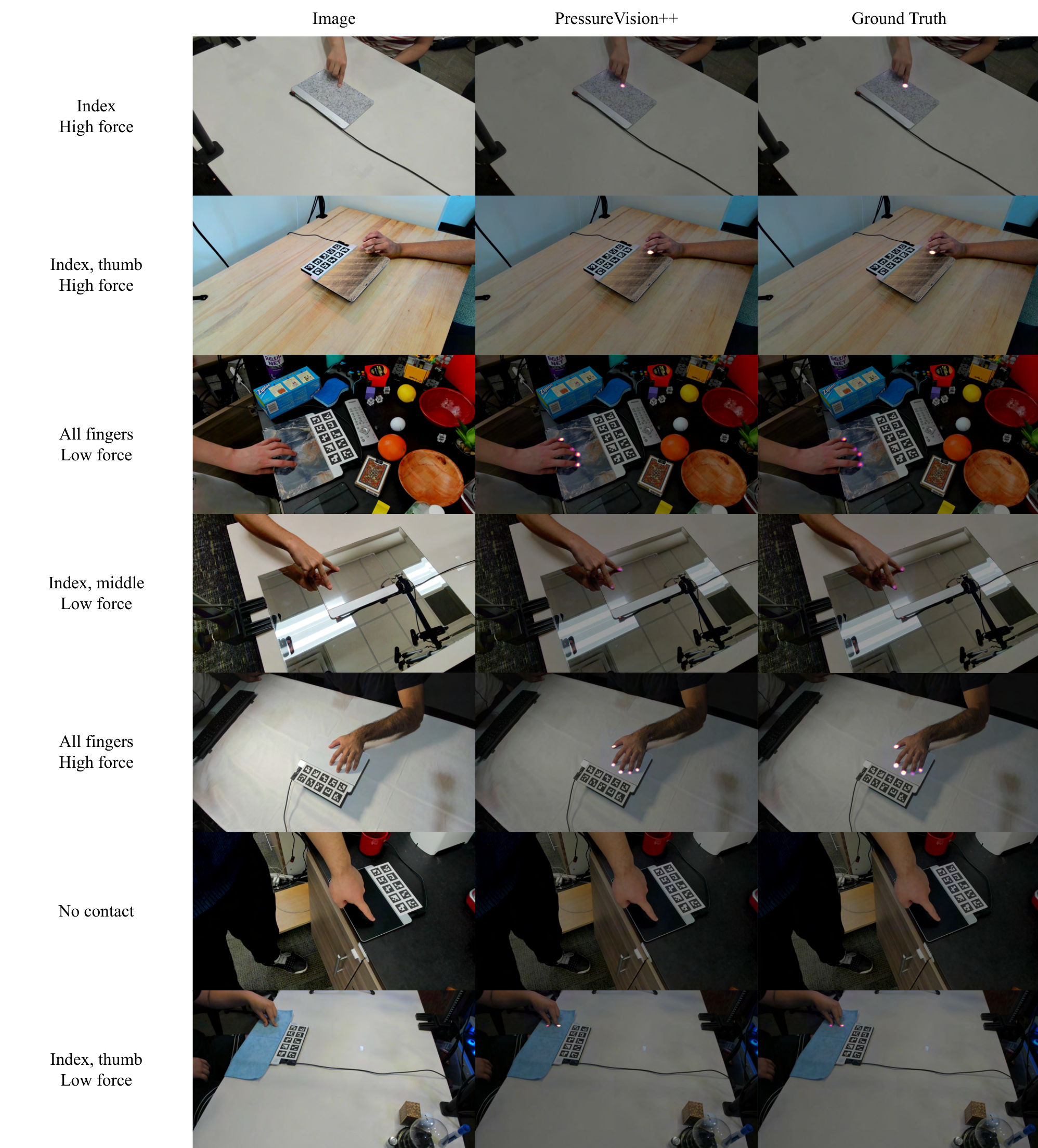}
  \caption{Results from the fully labeled test set where ground truth pressure is measured by a pressure sensor. Testing participants are held out from the training sets.}
  \label{fig:supp_collage_full_2}
\end{figure*}

\begin{figure*}
  \centering
  \includegraphics[width=0.65\linewidth]{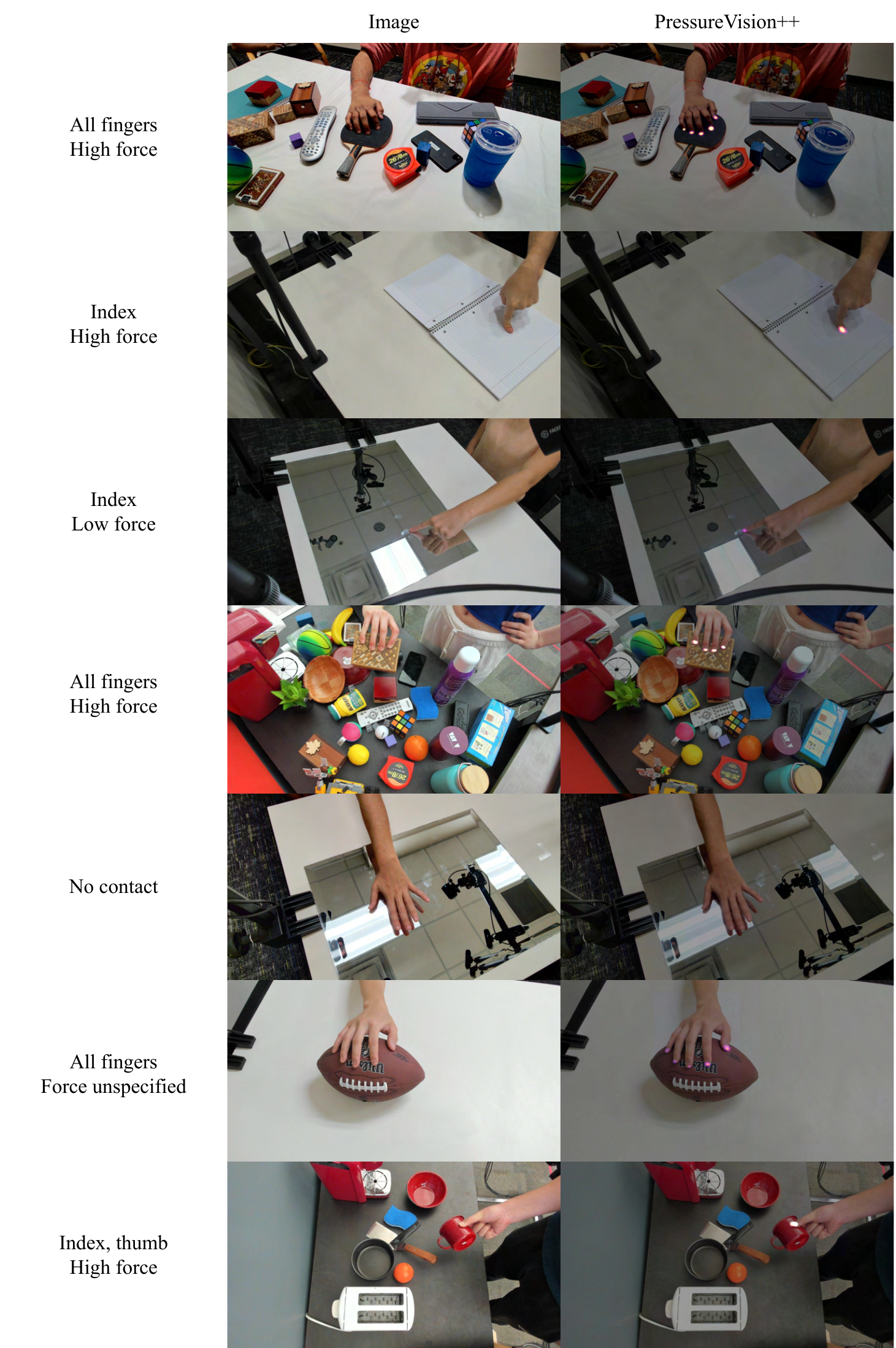}
  \caption{Results from the weakly labeled test set where no ground truth pressure is available. Testing participants are held out from the training sets.}
  \label{fig:supp_collage_weak_1}
\end{figure*}

\begin{figure*}
  \centering
  \includegraphics[width=0.65\linewidth]{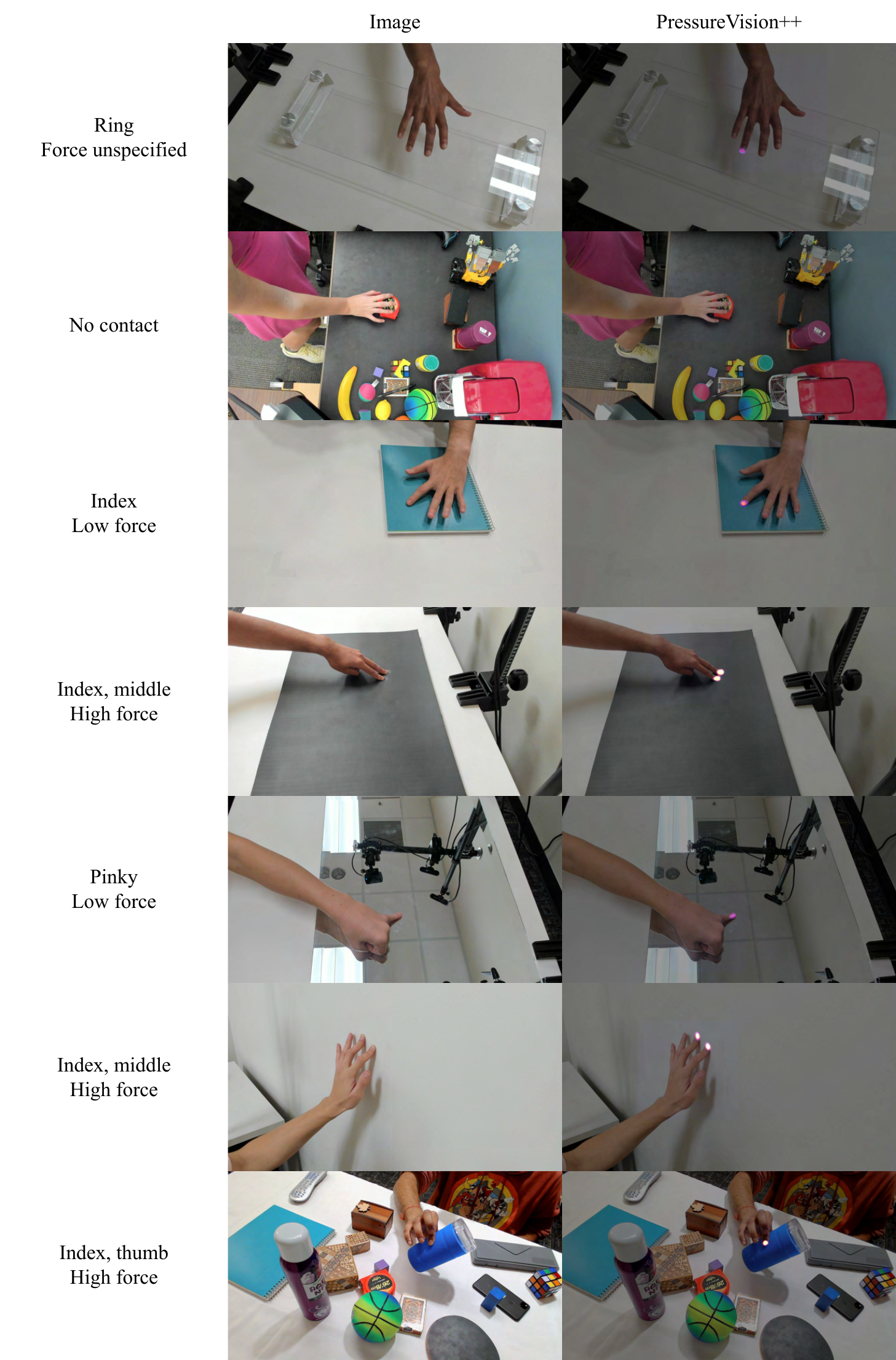}
  \caption{Results from the weakly labeled test set where no ground truth pressure is available. Testing participants are held out from the training sets.}
  \label{fig:supp_collage_weak_2}
\end{figure*}

In order to align coordinate frames between the RGB camera and the Meta Quest 2 headset, we designed a custom calibration tool (Figure \ref{fig:calibration_totem}). The calibration tool features an ArUco board \cite{garrido2014automatic} to estimate the pose of the RGB camera used for pressure estimation. The pose of the headset is calibrated by attaching a controller to the calibration tool. A calibration procedure is performed at the beginning of each session.

In order to calculate precise touch locations, the peaks of the pressure blobs are found with a local maxima detector. A custom application is developed for the Quest headset using Unity and the Oculus Integration Toolkit.

\subsection{Net WPM Metric}
For typing speed evaluations, words per minute (WPM) \cite{salthouse1984effects} is calculated by dividing the number of characters typed (including letters, spaces, and punctuation), $c$, by 5 to arrive at the number of words typed. Time $t$ is measured in seconds between the first keystroke and pressing ``Enter" to complete the sentence.
\begin{align}
    WPM=\frac{c/5}{t/60}
\end{align}

However, the WPM metric does not consider errors in typing. In our evaluations, we report net words per minute (Net WPM) \cite{salthouse1984effects}, which modifies the standard WPM metric to factor in errors. A single character error (insertion, deletion, or substitution) results in the subtraction of 5 characters, or one word. Where $e$ is the number of single-character errors, Net WPM can be calculated as:
\begin{align}
    Net WPM = \frac{c/5 - e}{t/60}
\end{align}

\subsection{Typing User Study}

For the typing user study, 10 participants were recruited who did not participate in the collection of \datasetName{} and who were not familiar with the research. The order of presentation of the two keyboards was randomized. Before collecting data, participants were allowed to practice typing with that keyboard for as long as they wanted.

After the study, participants were given a free-form text box to explain their perceived advantages and disadvantages of each typing method. They also rated which keyboard they preferred on a scale of 1 ``strongly prefer Direct Touch Keyboard" to 5 ``strongly prefer \methodName{} Keyboard". The average score was 4.4, with only one participant not preferring the \methodName{} keyboard.

We hypothesize as to the reasons why participants preferred the \methodName{} Keyboard. For the Direct Touch Keyboard, due to the noise in pose estimation, to prevent false keystrokes, participants must press each key very deeply. Additionally, users generally must look at their hands to find the correct key since it is difficult to memorize the location of mid-air keys. For the \methodName{} Keyboard, users only have to hover their fingers a few millimeters above the surface, and since the surface allows them to rest their hands and provides a reference point, they can type without looking at their hands. The most common error that participants made with the \methodName{} keyboard was pressing a key adjacent to the desired key, resulting in single-character errors. We hypothesize that a simple autocorrect system would be able to correct these errors easily and improve typing speed.

\begin{table*}
\centering
\begin{tabular}{l|c}
    \textbf{Action} & \textbf{Force Level}\\\hline
    Index, fingers & \{Low, high, slide, no contact\} \\\hline
    Thumb & \{Low, high, slide, no contact\} \\\hline
    Index and thumb & \{Low, high, slide, no contact\} \\\hline
    Index and middle & \{Low, high, slide, no contact\} \\\hline
    Middle & \{Low, high, slide, no contact\} \\\hline
    Ring & \{Low, high, slide, no contact\} \\\hline
    Pinky & \{Low, high, slide, no contact\} \\\hline
    All fingers & \{Low, high, slide, no contact\} \\\hline
    Press fingers sequentially & \{Low, high\} \\\hline
\end{tabular}
\caption{Participants were prompted according to this list of actions, e.g., \textit{press thumb, low force}. Not all participants completed the entire list of actions.}
\label{tab:action_prompts}
\end{table*}

\end{document}